\documentclass{article}

 \usepackage[nonatbib,final]{nips_2016}
\usepackage{cite}

\usepackage[T1]{fontenc}    
\usepackage{booktabs}       
\usepackage{amsfonts}       
\usepackage{nicefrac}       
\usepackage{microtype}      
\usepackage{epsfig}
\usepackage{graphicx}
\usepackage{amsmath}
\usepackage{amssymb}
\usepackage{epstopdf}
\usepackage{color,soul}

\title{Evolution in Groups: A deeper look at synaptic cluster driven evolution of deep neural networks}

\author{
  Mohammad Javad Shafiee$^{1,2}$,  Elnaz Barshan$^{1}$, Alexander Wong$^{1,2}$ \\
	$^{1}$Department of Systems Design Engineering, University of Waterloo, Waterloo, Ontario, Canada\\
$^{2}$DarwinAI, Waterloo, Ontario, Canada\\
	\texttt{\{mjshafiee, ebarshan, a28wong\}@uwaterloo.ca} \\
}
\begin{document}


\maketitle

\begin{abstract}
A promising paradigm for achieving highly efficient deep neural networks is the idea of \textit{evolutionary deep intelligence}, which mimics biological evolution processes to progressively synthesize more efficient networks. A crucial design factor in evolutionary deep intelligence is the genetic encoding scheme used to simulate heredity and determine the architectures of offspring networks. In this study, we take a deeper look at the notion of synaptic cluster-driven evolution of deep neural networks which guides the evolution process towards the formation of a highly sparse set of synaptic clusters in offspring networks. Utilizing a synaptic cluster-driven genetic encoding, the probabilistic encoding of synaptic traits considers not only individual synaptic properties but also inter-synaptic relationships within a deep neural network. This process results in highly sparse offspring networks which are particularly tailored for parallel computational devices such as GPUs and deep neural network accelerator chips. Comprehensive experimental results using four well-known deep neural network architectures (LeNet-5, AlexNet, ResNet-56, and DetectNet) on two different tasks (object categorization and object detection) demonstrate the efficiency of the proposed method. Cluster-driven genetic encoding scheme synthesizes networks that can achieve state-of-the-art performance with significantly smaller number of synapses than that of the original ancestor network. ($\sim$125-fold decrease in synapses for MNIST). Furthermore, the improved cluster efficiency in the generated offspring networks ($\sim$9.71-fold decrease in clusters for MNIST and a $\sim$8.16-fold decrease in clusters for KITTI) is particularly useful for accelerated performance on parallel computing hardware architectures such as those in GPUs and deep neural network accelerator chips.
\end{abstract}



\section{Introduction}
\vspace{-0.4 cm}
One of the key factors in revitalizing deep neural networks~\cite{lecun2015deep,graves2013speech,bengio2009learning,tompson2014joint} and their tremendous success is the significant growth in computational power. The proliferation of massively parallel computing devices such as graphics processing units (GPUs) and distributed computing has revolutionized the training and inference of deep neural networks due to their highly parallelizable nature. This incredible rise in the computational power enables researchers to design and build increasingly larger and deeper neural networks to boost modeling accuracy. This on-going growth in architectural complexity, however, has become a bottleneck in the widespread adoption of such deep neural networks in many operational scenarios. For example, in many applications such as self-driving cars, smart-phone applications, and surveillance cameras, the computational resources are limited to low-power embedded GPUs, CPUs and deep learning accelerator chips with strong constraints on the memory usage.  Furthermore, there are many situations where the use of cloud computing is intractable due to transmission cost, bandwidth issues, as well as privacy concerns.


Considering the obstacles associated with the architectural complexity of deep neural networks, we have witnessed a growing attention towards learning highly efficient deep neural network architectures that are able to provide strong modeling power for operational scenarios where limited memory and computational resources are available. One of the first approaches in this area is the optimal brain damage method~\cite{lecun1989optimal} in which synapses are pruned based on their strength. \mbox{Gong~{\it et al.}}~\cite{gong2014compressing} proposed a network compression framework where vector quantization is used to shrink the storage requirements of deep neural networks.  Han {\it et al.}~\cite{han2015learning} utilized Huffman coding in addition to pruning and vector quantization to further reduce the memory requirements. Hashing is another related trick employed by Chen {\it et al.}~\cite{chen2015compressing} for network compression. Low-rank approximation~\cite{ioannou2015training,jaderberg2014speeding,denton2014exploiting} and sparsity learning~\cite{feng2015learning,liu2015sparse,SSL_2016} are other strategies used to build smaller and more efficient deep neural networks.


Recently, Shafiee {\it et al.}~\cite{EvoNet1} introduced an evolutionary synthesis framework to progressively learn more efficient deep neural networks along successive generations\footnote{This framework is fundamentally different from the past attempts for utilizing evolutionary methods in training neural networks where genetic algorithm has been used to create neural networks~\cite{white1993gannet,stanley2002evolving,Angeline,Stanley,Gauci,Tirumala} with high modeling capabilities in a direct but highly computationally expensive manner.}. The proposed \textit{evolutionary deep intelligence} approach mimics biological evolution mechanisms such as random mutation, natural selection, and heredity within a probabilistic graphical modeling paradigm to successively synthesize more efficient offspring network architectures. A crucial design factor in evolutionary deep intelligence is the genetic encoding scheme used to simulate heredity and affects the architectural traits that are passed to the offspring networks in a significant way. Therefore, a more effective genetic encoding scheme can enable better transfer of genetic information from the ancestor network to its offspring networks to build a more efficient yet powerful future generation.


The introduced genetic encoding scheme in~\cite{EvoNet1} merely considers the individual synaptic properties in the sense that the probability of synthesizing each synapse within the network is independent of the rest of the synapses and thus it ignores the dependence between different synapses. However, there are neurobiological evidences that support the increasing probability of co-activation for \emph{correlated} synapses which encode similar information and locate close to each other on the same dendrite---synaptic clustering~\cite{welzel2010synapse, kastellakis2015synaptic, larkum2008synaptic, takahashi2012locally, winnubst2015synaptic}. Inspired by this observation, incorporating synaptic clustering in the genetic encoding scheme of evolutionary deep models is potentially a fruitful direction to investigate. Moreover, synthesizing the offspring networks based on synaptic clusters (instead of basing it purely on individual synapses) can increase the efficiency of the offspring deep neural networks running on parallel computing devices such as GPUs and deep neural network accelerator chips.

In this study, we take a deeper look at the notion of synaptic cluster-driven evolution of deep neural networks which guides the evolution process towards the formation of a highly sparse set of synaptic clusters in the offspring networks. This process results in highly sparse offspring networks which are particularly tailored for parallel computing devices such as GPUs and deep neural network accelerator chips. We introduce a multi-factor genetic encoding scheme in which the synaptic probability considers both the probability of synthesis for the cluster of synapses that includes a particular synapse and the probability of synthesis for that particular synapse within the cluster. This genetic encoding scheme effectively promotes the formation of synaptic clusters over successive generations during the evolution process while supporting the formation of highly efficient deep neural networks.


\section{Methodology }

Inspired by the neurobiological evidences, we propose a synaptic cluster-driven evolution framework for deep neural networks in which the ancestor network is guided towards the formation of a highly sparse set of synaptic \emph{clusters} in the offspring networks. The key idea here is to design a genetic encoding scheme that considers the \emph{inter-synaptic} relationships as well as the individual properties of each synapse. More formally, the synthesis probability distribution for each synapse is a product of the formation likelihood of the corresponding synaptic cluster and the synthesis probability of that particular synapse within that cluster.

\textbf{Evolutionary Deep Intelligence.}
Prior to describing the notion of synaptic cluster-driven evolutionary synthesis and the correspondingly proposed genetic encoding scheme, let us first provide an overview of the evolutionary deep intelligence framework introduced by Shafiee {\it et al.}~\cite{EvoNet1} for synthesizing progressively more efficient deep neural networks over successive generations within a probabilistic graphical modeling paradigm. In the evolutionary deep intelligence framework, the architectural traits of ancestor deep neural networks are encoded via probabilistic `DNA' sequences.  New offspring networks possessing diverse network architectures are then synthesized stochastically based on the `DNA' from the ancestor networks and probabilistic computational environmental factor models, thus mimicking random mutation, heredity, and natural selection. These offspring networks are then trained, much like one would train a newborn, and have more efficient, more diverse network architectures while achieving powerful modeling capabilities. A crucial design factor in evolutionary deep intelligence is the genetic encoding scheme used to mimic heredity, which can have a significant impact on the architecture of the evolved offspring networks.  In this study, we aim at exploring more effective genetic encoding schemes to guide the synthesis process so that the modeling capabilities of the ancestor network are faithfully captured by the more efficient offspring networks along successive generations.


\textbf{Cluster-driven Genetic Encoding.}
Let the network architecture of a deep neural network be expressed by $\mathcal{H}(N,S)$, with $N$ denoting the set of possible neurons and $S$ is the set of possible synapses in the network. Each neuron $n_i \in N$ is connected via a set of synapses $\bar{s} \subset S$ to neuron $n_j \in N$ such that the synaptic connectivity $s_i \in S$ is associated with a $w_i \in W$ denoting its strength.  Given the network architecture at the previous generation, i.e., $\mathcal{H}_{g-1}$, the architectural traits of a deep neural network in generation $g$, is encoded by the conditional probability $P(\mathcal{H}_g| \mathcal{H}_{g-1})$, which can be treated as the probabilistic `DNA' sequence of a deep neural network.

We assume that synaptic connectivity characteristics in an ancestor network are desirable traits to be inherited by the descendant networks. Therefore, the genetic information of a deep neural network is encoded in synaptic probability $P(S_g|W_{g-1})$, where $w_{k,g-1} \in W_{g-1}$ is the synaptic strength of each synapse $s_{k,g} \in S_{g}$.  In the proposed genetic encoding scheme for synaptic cluster-driven evolution, we aim at incorporating the neurobiological phenomenon of synaptic clustering~\cite{welzel2010synapse, kastellakis2015synaptic, larkum2008synaptic, takahashi2012locally, winnubst2015synaptic}, where the probability of synaptic co-activation increases for correlated synapses which encode similar information and are close together on the same dendrite.

To promote the formation of synaptic clusters over successive generations of more efficient offspring networks, we introduce a multi-factor synaptic probability model defined as follows:
\begin{align}
P(S_g|W_{g-1}) = \prod_{c \in C} \Big[P\big(\bar{s}_{g,c}|W_{g-1}\big) \cdot \prod_{i \in c }P(s_{g,i}|w_{g-1,i}) \Big]
\end{align}
where the first factor (first conditional probability) models the synthesis probability of a particular cluster of synapses, $\bar{s}_{g,c}$, while the second factor models the synthesis probability of a particular synapse, $s_{g,i}$, within synaptic cluster $c$.   More specifically, the probability $P(\bar{s}_{g,c}|W_{g-1})$ represents the likelihood that a particular synaptic cluster, $\bar{s}_{g,c}$, be synthesized as a part of the network architecture in generation $g$ given the synaptic strength in generation $g-1$.  For example, in a deep convolutional neural network, the synaptic cluster $c$ can be any subset of synapses such as a kernel or a set of kernels within the deep neural network.  The probability $P(s_{g,i}|w_{g-1,i})$ represents the likelihood of existence of synapse $i$ within the cluster $c$ in generation $g$ given its synaptic strength in generation $g-1$.  As such, the proposed synaptic probability model not only promotes the persistence of strong synaptic connectivity in offspring deep neural networks over successive generations, but also promotes the persistence of strong synaptic clusters in offspring deep neural networks over successive generations.
%

\textbf{Cluster-driven Evolutionary Synthesis.}  In the seminal paper on evolutionary deep intelligence by Shafiee~{\it et al.}~\cite{EvoNet1}, the synthesis probability $P(\mathcal{H}_g)$ is composed of the synaptic probability $P(S_g|W_{g-1})$, which mimic heredity, and environmental factor model $\mathcal{F}(\mathcal{E})$ which mimic natural selection by introducing quantitative environmental conditions that offspring networks must adapt to:
\begin{align}
P(\mathcal{H}_g) = \mathcal{F}(\mathcal{E}) \cdot P(S_g|W_{g-1})
\label{eq:EvoNet_EnvFactor}
\end{align}
In this study, \eqref{eq:EvoNet_EnvFactor} is reformulated in a more general way to enable the incorporation of different quantitative environmental factors over both the synthesis of synaptic clusters as well as each individual synapse, thus facilitating for synaptic cluster-driven evolution of deep neural networks:
\begin{align}
\label{eq:NewEvoNet_EnvFactor}
P(\mathcal{H}_g) =
\prod_{c \in C} \Big [\mathcal{F}_c (\mathcal{E}) P\big(\bar{s}_{g,c}|W_{g-1}\big) \cdot \prod_{i \in c} \mathcal{F}_s(\mathcal{E}) P(s_{g,i}|w_{g-1,i}) \Big]
\end{align}
where $\mathcal{F}_c (\cdot) $ and $\mathcal{F}_s(\cdot) $  represents environmental factors enforced at the cluster and synapse levels, respectively.

\textbf{Realization of Cluster-driven Genetic Encoding.}
To demonstrate the benefits of the proposed cluster-driven genetic encoding scheme, a simple realization of this scheme is presented in this study.  Here, since we wish to promote the persistence of strong synaptic clusters in offspring deep neural networks over successive generations, the synthesis probability of a particular cluster of synapses $\bar{s}_{g,c}$ is modeled as
\begin{align}
P\big(\bar{s}_{g,c} =1|W_{g-1}\big) = \exp \Big( \frac{\sum_{i \in c}\lfloor{\omega_{g-1,i}}\rfloor}{Z} - 1 \Big)
\label{eq:CLGE}
\end{align}
where $\lfloor{\cdot}\rfloor$ encodes the truncation of a synaptic weight and $Z$ is a normalization factor to make~\eqref{eq:CLGE} a probability distribution,  $P\big(\bar{s}_{g,c}|W_{g-1}\big)  \in [0,1]$.  The truncation of synaptic weights in the model reduces the influence of very weak synapses within a synaptic cluster on the genetic encoding process.  The probability of a particular synapse, $s_{g,i}$, within synaptic cluster $c$, denoted by $P(s_{g,i} = 1|w_{g-1,i})$ can be expressed as:
\begin{align}
P(s_{g,i} = 1|w_{g-1,i})  = \exp\Big( \frac{ \omega_{g-1,i}}{z} - 1 \Big)
\end{align}
where $z$ is a layer-wise normalization constant. By incorporating both of the aforementioned probabilities in the proposed scheme, the relationships amongst synapses as well as their individual synaptic strengths are taken into consideration in the genetic encoding process.
\section{Experimental Results}

 To study the efficacy of synaptic cluster-driven evolution for synthesizing highly efficient deep neural networks, evolutionary synthesis of deep neural networks across several generations was performed using the proposed genetic encoding scheme across four well-known deep neural network architectures for two different tasks:
 \begin{itemize}
 \item Object categorization using three benchmark datasets: MNIST~\cite{MNIST}, STL-10~\cite{STL10} and CIFAR10~\cite{CIFAR10}
 \item Object detection using one benchmark dataset: KITTI~\cite{KITTI}
 \end{itemize}

 For the MNIST and STL-10 experiments for object categorization, the LeNet-5 architecture~\cite{MNIST} is selected as the network architecture of the original, first generation ancestor network.  For the CIFAR10 experiment for object categorization, two different network architectures were explored as the first generation ancestor network.  First, the AlexNet architecture~\cite{krizhevsky2012imagenet} is utilized as one of the ancestor networks for CIFAR10, with the first layer modified to utilize $5 \times 5 \times 3$ kernels instead of $11 \times 11 \times 3$ kernels given the smaller image size in CIFAR10.  Second, the ResNet-56 architecture~\cite{resnet} is utilized as another ancestor network for CIFAR10, which allows us to study the behaviour of the proposed scheme for two very different deep neural network architectures.  For the KITTI experiment for object detection, the DetectNet architecture which is derived from GoogleNet~\cite{GoogleNet} that is tuned for object detection is selected as the network architecture of the original, first generation ancestor network.

The environmental factor model being imposed at different generations in this study is designed to form deep neural networks with progressively more efficient network architectures than its ancestor networks while maintaining modeling accuracy.
More specifically, $\mathcal{F}_c(\cdot)$ and $\mathcal{F}_s(\cdot)$ is formulated in this study such that an offspring deep neural network should not have more than 80\% of the total number of synapses in its direct ancestor network.
Furthermore, in this study, each kernel in the deep neural network is considered as a synaptic cluster in the synapse probability model.  In other words, the probability of the synthesis of a particular synaptic cluster (i.e, $P(\bar{s}_{g,c}|W_{g-1})$) is modeled as the truncated summation of the weights within a kernel.
%
%

\textbf{Architectural Efficiency Over Generations}. In this study, offspring deep neural networks were synthesized in successive generations until the accuracy of the offspring network exceeded 4\%, so that we can better study the changes in architectural efficiency in the descendant networks over multiple generations.  Table~\ref{Tab:Mnist-STL10Res} and Table~\ref{Tab:CIFAR10Res} show the architectural efficiency (defined in this study as the total number of synapses of the original, first-generation ancestor network divided by that of the current synthesized network) versus the modeling accuracy at several generations for three datasets and based on three different network architectures (LeNet-5, AlexNet, and ResNet-56). As observed in Table~\ref{Tab:Mnist-STL10Res}, the descendant network at the 13th generation for MNIST based on the LeNet-5 architecture was a staggering $\sim$125-fold more efficient than the original, first-generation ancestor network without exhibiting a significant drop in the test accuracy ($\sim$1.7\% drop).  This trend was consistent with that observed with the STL-10 results based on the LeNet-5 architecture, where the descendant network at the 10th generation was $\sim$56-fold more efficient than the original, first-generation ancestor network without a significant drop in test accuracy ($\sim$1.2\% drop). It also worth noting that since the training dataset of the STL-10 dataset is relatively small, the descendant networks at generations 2 to 8 actually achieved higher test accuracies when compared to the original, first-generation ancestor network, which illustrates the generalizability of the descendant networks compared to the original ancestor network as the descendant networks had fewer parameters to train.

For the case of CIFAR10 based on the AlexNet architecture, the descendant network at the 6th generation network was $\sim$14.4-fold more efficient than the original ancestor network with $\sim$2\% drop in test accuracy.  For the case of CIFAR10 based on the ResNet-56 architecture, the descendant network at the 7th generation network was $\sim$4-fold more efficient than the original ancestor network with $\sim$3\% drop in test accuracy.

Finally, Table~\ref{Tab:DetectNetA-E} shows the architectural efficiency versus modeling accuracy at several generations for the KITTI dataset and based on the DetectNet architecture.  For this case, the descendant network at the 11th generation was $\sim$10-fold more efficient compared to the original, first-generation ancestor network while achieving a $\sim$0.8\% increase in precision and a $\sim$1.7\% increase in recall.  These results not only demonstrate the efficacy of the proposed scheme, but also illustrate the applicability of the proposed scheme for a variety of different network architectures.

\begin{table}
	\caption{Architectural efficiency vs. test accuracy for different generations of synthesized networks for MNIST and STL-10 using LeNet-5. ``Gen.'', ``A-E'' and ``ACC.'' denote generation, architectural efficiency, and accuracy, respectively.  }
	\label{Tab:Mnist-STL10Res}
\begin{center}
\begin{minipage}{.5\textwidth}
\setlength\tabcolsep{0.15 cm}
\begin{center}
	\begin{tabular}{|c||c|c|}
	
				\multicolumn{3}{c}{\textbf{\footnotesize MNIST - LeNet-5}} \\\hline
			\footnotesize Gen.    & \footnotesize A-E & \footnotesize ACC.  \\ \hline \hline
			1   &1.00X         &99.47 \\
			5    & 5.20X  & 99.41 \\
			7  & 12.09X    &99.28\\
			11 &   62.74X    &98.49 \\
			13   &   125.09X & 97.75\\\hline
		\end{tabular}
\end{center}
\end{minipage}\hfill
\begin{minipage}{.5\textwidth}
 \setlength\tabcolsep{0.15 cm}
 \begin{center}
\begin{tabular}{|c||c|c|}
			\multicolumn{3}{c}{\textbf{\footnotesize STL-10 - LeNet-5}} \\\hline
			Gen.   & \footnotesize A-E &\footnotesize ACC.  \\ \hline \hline
			1  & 1.00X    & 57.74\\
			3 &  2.37X   &59.33\\
			7   &  14.99X   &60.51\\
			9  &   38.22X  &57.44\\
			10   &   56.27X  &56.58\\\hline
		\end{tabular}
\end{center}
\end{minipage}\hfill
\end{center}
\end{table}

\begin{table}
	\caption{Architectural efficiency vs. test accuracy for different generations of synthesized networks for CIFAR-10 using AlexNet and ResNet-56. ``Gen.'', ``A-E'' and ``ACC.'' denote generation, architectural efficiency, and accuracy, respectively.  }
	\label{Tab:CIFAR10Res}
\begin{center}
\begin{minipage}{.5\textwidth}
	\setlength\tabcolsep{0.15 cm}
\begin{center}
	\begin{tabular}{|c||c|c|}
		\multicolumn{3}{c}{\textbf{\footnotesize CIFAR10 - AlexNet}} \\\hline
		Gen.    &\footnotesize A-E &\footnotesize ACC.  \\ \hline \hline
		1    &  1.00X   & 86.69\\
		2    &   1.64X  & 88.14\\
		3   &  2.82X   &87.66 \\
		5   &   8.39X  & 85.88\\
		6   &   14.39X  & 84.59\\\hline

	\end{tabular}
\end{center}
\end{minipage}\hfill
\begin{minipage}{.5\textwidth}
	\setlength\tabcolsep{0.15 cm}
\begin{center}
	\begin{tabular}{|c||c|c|}
		\multicolumn{3}{c}{\textbf{\footnotesize CIFAR10 - ResNet-56}} \\\hline
		Gen.    &\footnotesize A-E &\footnotesize ACC.  \\ \hline \hline
		1    &  1.00X   & 93.00\\
		3   &  1.66X   &91.95 \\
		5   &  3.04X   &90.27 \\
		6   &  3.69X   &89.76 \\
		7   &  4.02X   &89.79 \\\hline

	\end{tabular}
\end{center}
\end{minipage}\hfill
\end{center}
\end{table}

\begin{table}
	\caption{Architectural efficiency vs. precision and recall for different generations of synthesized networks for KITTI using DetectNet. ``Gen.'', ``A-E'' denote generation and architectural efficiency, respectively.  }
	\label{Tab:DetectNetA-E}
		\setlength\tabcolsep{0.15 cm}
		\begin{center}
			\begin{tabular}{|c||c|c|c|}
				\multicolumn{4}{c}{\textbf{\footnotesize KITTI - DetectNet}} \\\hline
			\hline
				Gen.    &\footnotesize A-E  & \footnotesize Precision & \footnotesize Recall\\ \hline \hline
				1    &  1.00X   & 78.03 &62.03\\
				7   &   2.48X   & 81.93 &69.31\\
				10   &   5.24X  & 81.36 &67.57\\
				11   &   9.97X  & 78.84 &63.73\\
				\hline
				
			\end{tabular}
		\end{center}
\end{table}

\begin{table}[h]
	\caption{Cluster efficiency vs. test accuracy for the first and last reported generations of synthesized networks for MNIST and STL-10 using LeNet-5. ``Gen.'', ``C-E'' and ``ACC.'' denote generation, cluster efficiency, and accuracy, respectively.  }
	\label{Tab:Mnist-STL10ResC}
\begin{center}
\begin{minipage}{.5\textwidth}
\setlength\tabcolsep{0.15 cm}
\begin{center}
	\begin{tabular}{|c||c|c|}
	
				\multicolumn{3}{c}{\textbf{\footnotesize MNIST - LeNet-5}} \\\hline
			\footnotesize Gen.    & \footnotesize C-E & \footnotesize ACC.  \\ \hline \hline
			1   &1.00X         &99.47 \\
			13   &   9.71X & 97.75\\\hline
		\end{tabular}
\end{center}
\end{minipage}\hfill
\begin{minipage}{.5\textwidth}
 \setlength\tabcolsep{0.15 cm}
 \begin{center}
\begin{tabular}{|c||c|c|}
			\multicolumn{3}{c}{\textbf{\footnotesize STL-10 - LeNet-5}} \\\hline
			Gen.   & \footnotesize C-E &\footnotesize ACC.  \\ \hline \hline
			1  & 1.00X    & 57.74\\
			10   &   5.96X  &56.58\\\hline
		\end{tabular}
\end{center}
\end{minipage}\hfill
\end{center}
\end{table}

\begin{table}[h]
	\caption{Cluster efficiency vs. test accuracy for the first and last reported generations of synthesized networks for CIFAR-10 using AlexNet and ResNet-56. ``Gen.'', ``C-E'' and ``ACC.'' denote generation, cluster efficiency, and accuracy, respectively.  }
	\label{Tab:CIFAR10ResC}
\begin{center}
\begin{minipage}{.5\textwidth}
	\setlength\tabcolsep{0.15 cm}
\begin{center}
	\begin{tabular}{|c||c|c|}
		\multicolumn{3}{c}{\textbf{\footnotesize CIFAR10 - AlexNet}} \\\hline
		Gen.    &\footnotesize C-E &\footnotesize ACC.  \\ \hline \hline
		1    &  1.00X   & 86.69\\
		6   &   2.82X  & 84.59\\\hline

	\end{tabular}
\end{center}
\end{minipage}\hfill
\begin{minipage}{.5\textwidth}
	\setlength\tabcolsep{0.15 cm}
\begin{center}
	\begin{tabular}{|c||c|c|}
		\multicolumn{3}{c}{\textbf{\footnotesize CIFAR10 - ResNet-56}} \\\hline
		Gen.    &\footnotesize C-E &\footnotesize ACC.  \\ \hline \hline
		1    &  1.00X   & 93.00\\
		6   &   2.25X  & 89.79\\\hline

	\end{tabular}
\end{center}
\end{minipage}\hfill
\end{center}
\end{table}

\begin{table}[h]
	\caption{ Cluster efficiency vs. precision and recall for the first and last reported generations of synthesized networks for KITTI using DetectNet. ``Gen.'' and ``C-E'' denote generation and cluster efficiency, respectively.  }
	\label{Tab:Kitti}
	\begin{center}
		\begin{center}
			\begin{tabular}{|c||c|c|c|}
				\multicolumn{4}{c}{\textbf{\footnotesize KITTI - DetectNet}} \\\hline
			\hline
				Gen.    &\footnotesize C-E  & \footnotesize Precision & \footnotesize Recall\\ \hline \hline
				1    &  1.00X   & 78.03 &62.03\\
				11   &   8.16X  & 78.84 &63.73\\
				\hline
				
			\end{tabular}
		\end{center}
	\end{center}
\end{table}

\textbf{Cluster Efficiency}. Table~\ref{Tab:Mnist-STL10ResC} and~\ref{Tab:CIFAR10ResC} shows the cluster efficiency of the synthesized deep neural networks in the last generations, where cluster efficiency is defined in this study as the total number of kernels in a layer of the original, first-generation ancestor network divided by that of the current synthesized network. It can be observed that for MNIST, the cluster efficiency of the last-generation descendant network is $\sim$9.7X, which may result in a near 9.7-fold potential speed-up in running time on embedded GPUs and deep neural network accelerator chips by reducing the number of arithmetic operations by $\sim$9.7-fold compared to the first-generation ancestor network, though computational overhead in other layers such as ReLU may lead to a reduction in actual speed-up.  The potential speed-up from the last-generation descendant network for STL-10 is lower compared to MNIST dataset, with the reported cluster efficiency in last-generation descendant network  $\sim$6X.

The cluster efficiency for the last generation descendant networks for CIFAR10 using AlexNet and ResNet-56 are $\sim$2.8X and $\sim$2.25X, respectively.  Finally, Table~\ref{Tab:Kitti} demonstrates the cluster efficiency of the synthesized deep neural networks in the last generations for the KITTI dataset using DetectNet. As seen, the cluster efficiency of the last generation descendent network is $\sim$8.16X, while achieving a $\sim$0.8\% increase in precision and a $\sim$1.7\% increase in recall compared to the original, first-generation ancestor network. These results demonstrate that not only can the proposed genetic encoding scheme promotes the synthesis of deep neural networks that are highly efficient yet maintains modeling accuracy, but also promotes the formation of highly sparse synaptic clusters that make them highly tailored for devices designed for highly parallel computations such as GPUs and deep neural network accelerator chips.

\vspace{-0.25 cm}
\section{Conclusion}
\vspace{-0.4 cm}

In this study, we took a deeper look at a synaptic cluster-driven strategy for the evolution of deep neural networks, where synaptic probability within a deep neural network is driven towards the formation of highly sparse synaptic clusters.  Experimental results for the tasks of object categorization and object detection demonstrated that the synthesized `evolved' offspring networks using this evolution strategy via a synaptic cluster-driven genetic encoding scheme can achieve state-of-the-art performance while having network architectures that are not only significantly more efficient compared to the original ancestor network, but also highly tailored for operational machine learning applications and scenarios using devices designed for highly parallel computations such as GPUs and deep neural network accelerator chips.
Future work involves investigating alternative realizations of this new genetic encoding scheme beyond the simple realization presented in this study, and study the network architecture evolutions of deep neural networks synthesized by these realizations over successive generations to better understand their effectiveness and efficiency.
\vspace{-0.05 cm}
\subsubsection*{Acknowledgments}
This research has been supported by Natural Sciences and Engineering Research Council of Canada (NSERC) and the Canada Research Chairs program.  The authors also thank Nvidia for the GPU hardware used in this study through the Nvidia Hardware Grant Program.






\end{document}